\documentclass{amsart}

\usepackage[utf8]{inputenc}
 \usepackage{fullpage}
 \usepackage[hidelinks]{hyperref}
\usepackage{listings}
 \usepackage{color}
\usepackage[dvipsnames]{xcolor}

\usepackage{amsmath}
\usepackage{amsfonts}
\usepackage{amssymb}
\usepackage{amsthm}
\usepackage{mathtools}
\usepackage{dsfont}
\usepackage{booktabs}
\usepackage{subcaption}
\usepackage[capitalise]{cleveref}
\Crefname{subsection}{Subsection}{Subsections}
\crefname{subsection}{Subsection}{Subsections}
\crefname{line}{Line}{Lines}
\usepackage{algorithm}
\usepackage[noend]{algpseudocode}
\usepackage{mathrsfs}
\usepackage{comment}
\usepackage{chronology}
\usepackage{float}
\usepackage{wrapfig}
\usepackage{xspace}

\usepackage[utf8]{inputenc}
\usepackage{fourier} 
\usepackage{array}
\usepackage{makecell}

\newcommand{\R}{\mathbb{R}} 
 
\newcommand{\E}{\mathbb{E}}

\newcommand{\B}[1]{\mathbf{#1}}

\theoremstyle{plain}

\theoremstyle{definition}

\theoremstyle{remark}

\DeclareMathOperator{\sgn}{sign}

\newcommand{\BF}[1]{\mathbf{#1}}
\newcommand{\BB}[1]{\mathbb{#1}}

\newcommand{\ie}[0]{\textit{i.e.}\xspace}
\newcommand{\eg}[0]{\textit{e.g.}\xspace}

\newcommand{\iid}[0]{\textit{i.i.d.}\xspace}

\newcommand{\TODO}[1]{}

\definecolor{mypink3}{cmyk}{0, 0.7808, 0.4429, 0.1412}

\title{Nonlinear Matrix Approximation with Radial Basis Function Components}

\author{
  Elizaveta Rebrova and Yu-Hang Tang\\  
  Computational Research Division\\
  Lawrence Berkeley National Laboratory\\
  Berkeley, CA 94598, USA \\
  \texttt{elizaveta.rebrova@gmail.com, tang.maxin@gmail.com} \\
}

\begin{document}

\begin{abstract}
    We introduce and investigate matrix approximation by decomposition into a sum of radial basis function (RBF) components.
    An RBF component is a generalization of the outer product between a pair of vectors, where an RBF function replaces the scalar multiplication between individual vector elements.
    Even though the RBF functions are positive definite, the summation across components is not restricted to convex combinations and allows us to compute the decomposition for \emph{any} real matrix that is not necessarily symmetric or positive definite. 
    We formulate the problem of seeking such a decomposition as an optimization problem with a nonlinear and non-convex loss function.
    Several modern versions of the gradient descent method, including their scalable stochastic counterparts, are used to solve this problem.
    We provide extensive empirical evidence of the effectiveness of the RBF decomposition and that of the gradient-based fitting algorithm.
    While being conceptually motivated by singular value decomposition (SVD), our proposed nonlinear counterpart outperforms SVD by drastically reducing the memory required to approximate a data matrix with the same $L_2$-error for a wide range of matrix types. For example, it leads to 2 to 6 times memory save for Gaussian noise, graph adjacency matrices, and kernel matrices.
     Moreover, this proximity-based decomposition can offer additional interpretability in applications that involve, e.g., capturing the inner low-dimensional structure of the data, retaining graph connectivity structure, and preserving the acutance of images.
\end{abstract}

\maketitle

\section{Introduction}

There is an extensive body of work that aims to approximate large multi-dimensional data via simple low-dimensional components. Such approximations save the storage memory and expose simpler underlying structures in the data. The singular value decomposition (SVD) for matrices is arguably the standard example of such a low-rank approximation, while the canonical polyadic (CP) decomposition~\cite{carrollAnalysisIndividualDifferences1970,harshmanFoundationsPARAFACProcedure1970a}, where every component is represented by the outer product of two or more vectors, is the generalization for higher-order tensors. Much less is known for approximations that are low-complexity but not necessarily low-rank. In this work, we propose one such approximation by generalizing the outer product between vectors. By \emph{generalizing}, we mean to replace the scalar product between vector elements by a nonlinear function, \eg the one-dimensional radial basis function (RBF) in this case. Just like a component formed by an outer product, each nonlinear component is indexed by two vectors with potentially one additional scaling factor. However, the nonlinear relationship introduced between the vectors leads to more expressive components as manifested by, for example, their higher rank. This in turn yields better overall approximations with less explicit components comparing to the SVD in many situations. 

The generalization of the outer product using RBF functions stems from their use in kernel machines~\cite{hofmannKernelMethodsMachine2008}. In that context, a kernel is a positive definite function that evaluates the inner product between data points after implicitly mapping them to a higher-dimensional vector space. Specifically, the RBF kernel between two feature vectors $\B{u}, \B{v} \in \mathbb{R}^n$ is a real-valued function of the form
\begin{equation}\label{eq:rbf-kernel}
k_\text{RBF}(\B{u}, \B{v}) = \exp\left(-\frac{\|\B{u} - \B{v}\|^2}{2h^2}\right),
\end{equation}
where $h$ is a scalar parameter. The mapping, which is nonlinear, introduces flexibility for modeling functions and decision boundaries that would otherwise be difficult to learn in the original feature space. Moreover, the RBF kernel can be conveniently interpreted as a similarity measure, \ie  the value of the kernel is small if input vectors are far away from each other.

In our work, an \emph{RBF component} created from two vectors $\B{u} = (u_1, \ldots, u_n) \in \mathbb{R}^n$, $\B{v} = (v_1, \ldots, v_m) \in \mathbb{R}^m$ is a kernel matrix $K = \left(k_{ij}\right) \in \mathbb{R}^{n \times m}$ with the elements defined as
\begin{equation}\label{eq:rbf-comp}
    k_{ij} = \exp\left( -\left|u_i - v_j\right|^2 \right).
\end{equation}
 An important difference here, as opposed to the case in \eqref{eq:rbf-kernel}, is that the vectors $\B{u}, \B{v}$ are not known inputs but rather \emph{learnable} parameters for expressing a given target data matrix. In particular, there is no need to specify the length-scale parameter $h$ as in the standard RBF definition~\eqref{eq:rbf-kernel}, since optimizing $\B{u}$ and $\B{v}$ can achieve an equivalent effect.  Similarly to the RBF function~\eqref{eq:rbf-kernel}, an RBF component~\eqref{eq:rbf-comp} exposes the underlying geometric structures in the data matrices based on the proximity between the elements of the learned vectors.
 However, we emphasize that the approximations are not specific to any special geometry-related features in the data. Linear combinations of such kernel matrices can act as universal approximators of general matrices while competing successfully in efficiency with SVD decomposition.

The task of finding and interpreting an optimal RBF approximation results in unique challenges distinct from SVD-based approximation algorithms. This includes the absence of known direct algorithms to find the best decomposition, the lack of orthogonality between the individual components, and the lack of a clear ordering of the components based on their norm. However, we show that trading-in the linearity of the components has great advantages in memory efficiency and additional expressivity of the decomposition due to the Euclidean distance in the structure of the RBF kernels. We also demonstrate that iterative gradient-based optimization methods make the problem of finding a good approximation tractable. 

This paper is structured as follows. In Section~\ref{sec:motivating-example}, we present a synthetic case where an RBF approximation is naturally optimal because the matrix to be approximated is created from the difference between two RBF components. This sets the scene for a formal description of our approximation method, including a discussion of the success and sophistication of the learning process, as presented in Section~\ref{sec:method}. In Section~\ref{sec:efficiency}, we provide extensive empirical evidence of the general applicability of the method to a variety of natural matrix models without \textit{a priori} underlying structures that make them particularly amenable to the RBF approximation. In Section~\ref{sec:interpretability}, we discuss how additional knowledge about the structure in data can be obtained using the learned RBF components in several situations, including a manifold learning example, a graph structure learning example, and an image compression example. In \Cref{sec:related}, we compare the proposed approach with related works.
In \Cref{sec:conclusions}, we conclude the paper with a discussion of multiple future directions related to the analysis and applications of the proposed RBF approximation.

\section{A Motivating Example}\label{sec:motivating-example}

Let us consider a special case when an RBF component defined in \eqref{eq:rbf-comp} is square and symmetric, namely, $\bf{u} = \bf{v}$ and
\begin{equation}\label{eq:rbf-component-sym}
\tilde k_{ij} \coloneqq \exp\left[-(u_i - u_j)^2\right], \quad \text{for each entry } \tilde k_{ij} \text{ in } \tilde K \in \mathbb{R}^{n \times n}.
\end{equation}  Depending on the vector $\B{u}$, a symmetric RBF component can have vastly different structures from a rank one constant matrix to a full-rank near identity matrix. In the case where $\B{u}$ is randomly sampled from a regular enough distribution, the spectrum of the normalized matrix  $n^{-1}K$ is concentrated around the spectrum of its expectation $n^{-1}\E K$~\cite{koltchinskii2000random}. For example, when the vector elements are \iid standard Gaussians, with high probability, the RBF component has $9$ eigenvalues above $10^{-5}$.

\begin{figure}
    \includegraphics[width=\textwidth]{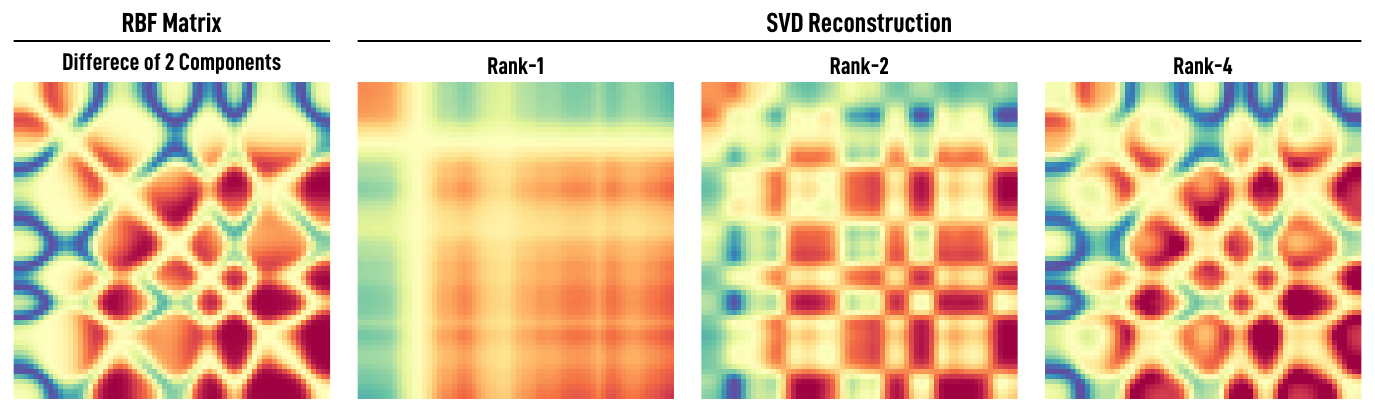}
	\caption{The best rank 1, 2 and 4 approximations of $K_{exact2}$ matrix are visibly not ideal, even though the latter one requires twice more parameters than the exact functional expression of $K_{exact2} = 5K^{(1)} - 4K^{(2)}$ where the matrices $K^{(i)}$ are given by the form \eqref{eq:rbf-component-sym}.}
    \label{fig:motivating}
\end{figure}

A fact that we take as an essential advantage for approximating general matrices is that a linear combination of two or more RBF components is not necessarily semi-positive definite. Figure~\ref{fig:motivating} shows one such example of a two-component matrix $K_{exact2} = 5K^{(1)} - 4K^{(2)}$, where $K^{(1)}$ and $K^{(2)}$ are the RBF components defined per \eqref{eq:rbf-component-sym}. Without loss of generality, we take $\B{u}^{(1)}$ and $\B{u}^{(2)}$ to be standard normal vectors smoothed by one-dimensional Gaussian filters with standard deviations of 3 and 6, respectively. Even though the matrix can be perfectly reconstructed from only $2n$ parameters that encode $\B{u}^{(1)}$ and $\B{u}^{(2)}$, the best rank-4 SVD approximation has a mean square error above $0.01$ and delivers a visibly imperfect reconstruction using $4n$ parameters.

\begin{figure}
    \includegraphics[width=\textwidth]{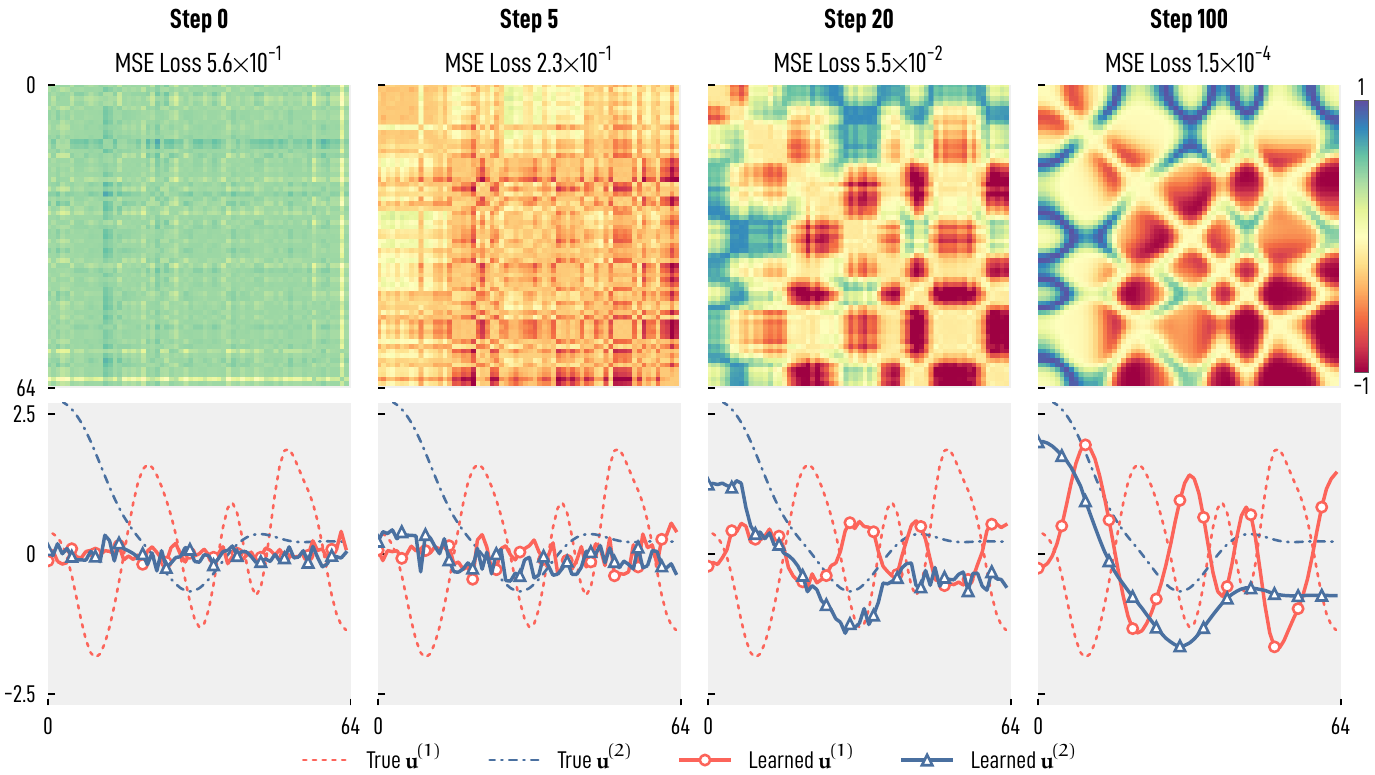}    \caption{Learning process via Adam method of the two RBF component approximation of $K_{exact2}$ matrix: initial matrix is nearly constant (first row), since randomly initialized component vectors $\B{u^{(1)}}$ and $\B{u^{(2)}}$ are chosen to have small coordinates all about 1e-2 (second row), and, as the optimization proceeds, after 5, 20 and 100 steps, the component vectors gradually learn the shape of the true component vectors, and the learned matrix approximates $K_{exact2}$.}
    \label{fig:iteration}
\end{figure}

In this paper, we are motivated by the following \emph{inverse} questions: Would this be possible to recover vectors $\B{u}^{(1)}$ and $\B{u}^{(2)}$ if we were just given a matrix $K_{exact2}$, similar to how the SVD reconstruction in \Cref{fig:motivating} was found? More generally, given a matrix $K$ that is known to be equal to, or well-approximated by a linear combination of $r$ general RBF components as per \eqref{eq:rbf-comp}, can we learn the vectors ${\bf u}^{(k)}, \B{v}^{(k)}$ and scalar coefficients that define this linear combination? Further, can we find a linear combination of the RBF components to approximate a generic data matrix $K$ with no initial structure assumed? This is parallel to finding an SVD approximation of a given rank for a target matrix but poses a much more challenging optimization problem due to the nonlinearity of the RBF components. In this paper, we give promising answers to all three key questions above. In Figure~\ref{fig:iteration}, we demonstrate that for a matrix $K_{exact2}$, the component vectors $\B{u}^{(1)}$ and $\B{u}^{(2)}$ can be found by an efficient iterative optimization process. In the next section, we dive into the specifics of this training process that allows us to learn the RBF component vectors and coefficients for a given data matrix.

\section{The Optimization Problem}\label{sec:method}
For a matrix $K =(k_{ij}) \in \BB{R}^{n\times m}$, constructing an $r$-component RBF approximation is a problem of finding a sequence of column vectors $\{\BF{u}^{(k)} \in \BB{R}^n, k = 1, \ldots, r\}$ and row vectors $\{\BF{v}^{(k)} \in \BB{R}^m, k = \{1, \ldots, r\}$, a vector of coefficients $\BF{a} \in \BB{R}^r$ and a scalar offset $b$ such that
\begin{equation}
    k_{ij} \approx k_{ij}^0 := b + \sum_{k=1}^r a_k \exp \left[- \left(u_i^{(k)} - v_j^{(k)} \right)^2 \right]
\end{equation}

This leads to the following optimization problem
\begin{equation}
    \underset{\{\BF{u}^{(k)}\}, \{\BF{v}^{(k)}\}, \BF{a}, b}{\mathrm{argmin}} \mathcal{L}(K, K^0), \quad \text{ where } \ \mathcal{L}(K, K^0) = \frac{1}{nm}\sum_{i,j =1, 1}^{n,m} \left(k_{ij} - k_{ij}^0 \right)^2
 \label{eq:optimization-problem}
\end{equation}
is the mean squared error loss function. Although other losses could be considered depending on an application, we choose an $L_2$-based loss to establish direct comparison with SVD-based approximations, which is the best low-rank approximation for the loss as guaranteed by the Eckart–Young–Mirsky theorem \cite{eckart1936approximation}.

\textbf{Global minimum structure.} The problem \eqref{eq:optimization-problem} is nonlinear, non-convex, and potentially of very high dimension $\sim \mathcal{O}\left((n+m) k\right)$. In particular, the global minimum may not be unique as multiple equivalent solutions exist due to the permutation of the RBF components. It is only obvious in the one-component case, where the entries of $K$ correspond to one-dimensional proximities between the vector elements as discussed above, that the set of global minima consists of the ground truth vector $\B{u}$ uniquely determined up to reflection and a constant shift. Empirically, we can see in~\Cref{fig:iteration} how two learned component vectors differ from true component vectors only by isometries.

\begin{figure}
    \includegraphics[width=\textwidth]{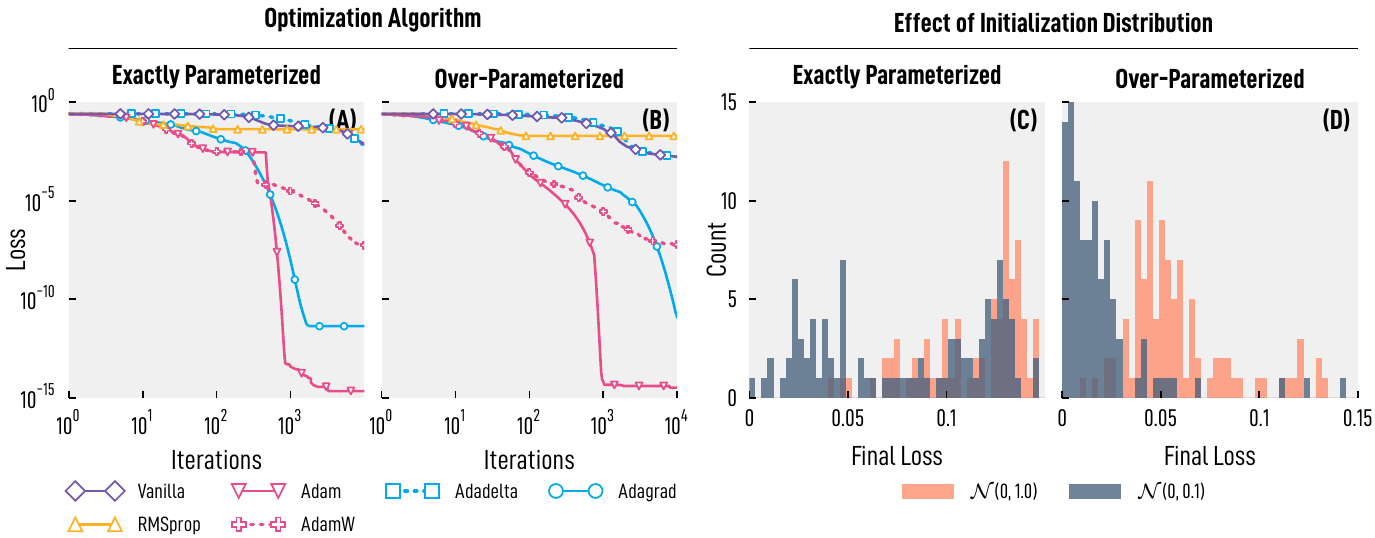}    \caption{Learning two (exactly-parametrized regime) and four (over-parametrized regime) RBF component approximations of $K_{exact2}$: (A) and (B) show that standard adaptive gradient-based learning methods, such as Adam, AdamW and Adagrad, can find approximation $K_{exact2}$ with the MSE loss below 1e-5 in about $1000$ iterations of $100$ independent runs, and that convergence tends to be quicker in the over-parametrized regime. (C) and (D) show the distribution of the resulting MSE approximation loss over $100$ independent randomly started runs after $10K$ iteration steps of the Adam algorithm, depending on the starting random distribution of the component vector coefficients (standard Gaussian, or 10 times damped). The initialization with the smaller random values, and especially the over-parametrization, lead to additional robustness of the learning process.}
    \label{fig:optimization}
\end{figure}

\textbf{Adaptive iterative gradient-based optimization: Adam.} We turn to \emph{gradient-based optimization techniques} for solving the problem~\eqref{eq:optimization-problem}. It is known that the stochastic gradient descent method used with adaptive step size algorithms can successfully handle complex non-convex objectives~\cite{wilson2017marginal, o2019behavior, de2018convergence}. Their practical use in the deep network community goes way beyond their established theoretical guarantees. As visualized in~\Cref{fig:optimization}, the learning dynamic of the two-component matrix $K$, introduced in Section~\ref{sec:motivating-example}, suggests that Adam~\cite{kingma2014adam} can generally outperform other popular gradient descent algorithms. Therefore, we use it for all subsequent experiments presented in the paper.

\textbf{Vanishing gradients and small magnitude initialization.} The problem of vanishing gradient may halt the training process when entries of the learned vectors have large mutual distances. To illustrate this issue, let us consider the problem of learning the vector $\B{u}$ from a matrix $K$ of the simplest one-component symmetric form \eqref{eq:rbf-component-sym}. The corresponding loss function is
\begin{equation}\label{eq:one-component-loss}
\mathcal{L} = \frac{1}{n^2}\sum_{ij} \left[\exp(-(u_i - u_j)^2) - k_{ij} \right]^2
\end{equation}
and its $(i,j)$-th gradient is given by $-2/n^2(u_i - u_j)\exp(-(u_i - u_j)^2)$, which approaches zero both for very large and especially small values of $|u_i - u_j|$. In particular, the training stalls (or stops) if we start with the points that are too far from each other. We partially alleviate this issue by making sure to initialize the parameters in component vectors $\B{u}$ with small enough values. The experimental evidence confirms that this makes the training process more consistent: the histograms in \Cref{fig:optimization} show the distributions of the losses achieved on the batch of $100$ runs, all started randomly with the independent standard normal entries, multiplied by a factor of $1$ and $0.1$ respectively. We can see that smaller initialization values lead to a larger fraction of independent training runs resulting in a global minimum. We can also see in~\Cref{fig:iteration} how the magnitude of the parameters gradually grows during the learning process.

\textbf{Batch training: only the best-run matters.} However, the histograms in \Cref{fig:optimization} show a non-negligible fraction of runs that do not converge to a global minimum, for both initialization options. The crucial advantage of our problem comparing to the majority of typical machine learning problems (when a gradient-based approach is used for training the model) is that we always know the ground truth matrix $K$, and we do not need the learned parameters to generalize to the unseen data. Thus, we can track the error during the training process. The following process can be employed to get a consistent convergence result: after training a batch of gradient descent processes with various random starts, we keep only the best one (or several), and restart all the others. In this paper, we employ a simpler approach: we simply run a \emph{batch of independent learning processes} concurrently and remember that it is enough to get just one good run to learn an efficient RBF approximation. Essentially, our task requires an algorithm design that will result in a non-negligible (and not necessarily high) probability of successful training.

\textbf{Random initialization and over-parametrization.} It is known that the gradient-based methods are very sensitive to the initialization, and it is always advantageous to have a warm start for the training. Although the magnitudes of the entries of $K$ give some information about the distances between the coordinates of the component vectors, such effect is hard to quantify especially for fitting an approximation with multiple RBF components. Here, we do not tune the starting distribution for the specific data matrix, but rather initialize randomly. The promise of random initialization was proposed in several recent papers, including~\cite{wang2020beyond}, where the authors demonstrate that random initialization works for a tensor CP-fitting problem in the over-parametrized regime. In our RBF fitting problem, over-parametrization also leads to the increased speed and robustness of the training process, as we can see on the right two displays in \Cref{fig:optimization}.

\textbf{Stochastic counterparts.} \emph{Stochastic} gradient descent is often used in training deep neural networks to reduce the computational cost while also providing the ability to overcome local minima as a side effect of the introduced randomization. The key here is to estimate the gradient of the loss function with respect to the learnable parameters using a small subset of the training examples chosen randomly during each gradient descent iteration. In the context of solving~\eqref{eq:optimization-problem} for the RBF decomposition, this translates into evaluating the loss function over a subset of the elements of the target matrix. In other words, we evaluate the loss function as
\begin{equation}
    \tilde{\mathcal{L}} = \frac{1}{|I|}\sum_{i, j \in I} (k_{ij}- k^0_{ij})^2,
\end{equation}
where $I$ is an index set that contains a random subset of the elements of the matrix coordinates and $|I|$ denotes its cardinality.

Let us conclude this section with a disclaimer that it is still an important direction for future work to analyze the loss landscape specific to the RBF fitting problem and to develop tailored initialization and optimization processes for this particular shape of the loss function.

\section{Efficiency of the RBF approximation}\label{sec:efficiency}
As we discussed in Section~\ref{sec:motivating-example}, an RBF approximation is more efficient than any low-rank approximation for at least \emph{some} matrices. Further, observing that \emph{every} matrix is exactly a linear combination of several rank-one matrices, we claim that a natural analog holds: every matrix can be approximated by a linear combination of the RBF components given by ~\eqref{eq:rbf-comp}. One can notice that by constructing $u_{i'} = v_{j'}$ for a fixed pair of indices $(i',j')$ and setting all other elements very large so that $|u_i - v_j|, i \neq i' \text{or} j \neq j'$ is greater than multiple length scales of the RBF function, we obtain a matrix $\tilde K$ very close to being having exactly one non-zero entry. This gives a trivial, albeit memory-heavy, universal approximation for any matrix in $\R^{n \times m}$ using at most $nm$ components of type \eqref{eq:rbf-comp}. Of course, approximating only one matrix element using an RBF component is completely impractical, since storing a truncated SVD decomposition or even storing the elements of the original matrix in verbatim would take less memory. This motivates the next key question we address in this paper: \emph{For what natural models is the RBF approximation of a general data matrix K more successful than one based on SVD?} 

In this section, we compare the efficiency of the approximations by RBF and SVD components in terms of reconstruction quality as measured by the MSE loss~\eqref{eq:optimization-problem} over a variety of the data matrix models.  Note that in the symmetric case, right and left singular vectors coincide, and so the SVD-based low-rank approximation depends on twice fewer parameters. Parallel to this, we use symmetric RBF components of form~\eqref{eq:rbf-component-sym} to approximate symmetric matrices. Finally, it is worth noting that there are ``adversarial" settings where SVD approximations will show better results than the RBF ones. For example, a rank $1$ matrix can be approximated exactly by one SVD component, but generally not by one RBF component. However, as the rank $r$ of the target matrix grows, we can see that the rank $r' \lesssim r$ approximations by the RBF components quickly become more efficient. See also the discussion about the correct functional form below in Section~\ref{sec:conclusions}.

\subsection{Synthetic Data}

First, we consider the following artificial random matrix models: 1) a square matrix of random Gaussian noise, 2) a rectangular matrix of random Gaussian noise, 3) the adjacency matrix of the Erd\H{o}s-R\'{e}nyi graph, and 4) the adjacency matrix of the Barab\'{a}si-Albert graph~\cite{barabasi1999emergence}. We show that RBF component approximations are more efficient than truncated SVD approximations even for these highly unstructured models.
This suggests that the RBF approximation is not merely more efficient due to the exploitation of special structures in the data, \eg as exemplified in \Cref{fig:motivating}. The RBF components learn \emph{general} data matrices better while being defined by the same number of parameters as the standard outer products because they are more informative and flexible as manifested by the higher number of ranks that each component possesses.

\begin{figure}
    \includegraphics[width=\textwidth]{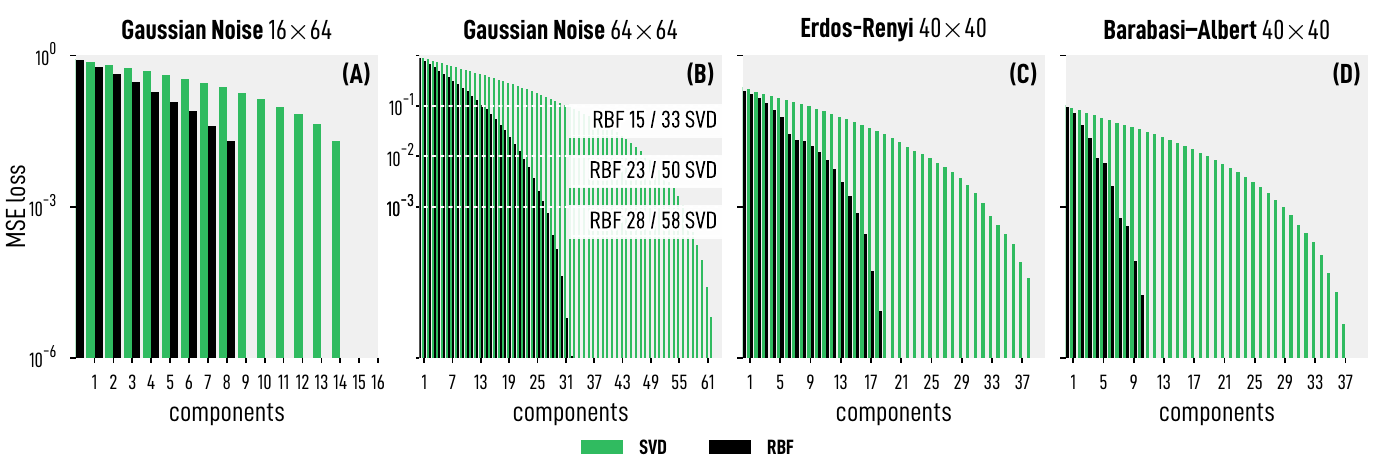}
	\caption{A comparison of MSE error achieved by RBF and SVD approximations depending on the number of components/ranks. Various full-rank random models considered: (A) rectangular and (B) square standard asymmetric Gaussian matrices and (C) and (D) adjacency matrices of random graphs. Consistenly over these models, several times less parameters are needed to achieve the same MSE accuracy with the RBF approximation.}
    \label{fig:random_efficiency}
\end{figure}

\Cref{fig:random_efficiency} (A)/(B) shows the decay of the mean square loss as we increase the number $r$ of RBF components. For the random matrices with independent standard Gaussian entries, approximately twice the number of SVD ranks are required to achieve the same mean square error with the RBF approximations.

\Cref{fig:random_efficiency}(C)/(D) show the decay rate of reconstruction loss for the adjacency matrices of a Erd\H{o}s-R\'{e}nyi random graph and a Barab\'{a}si-Albert random graph. The Erd\H{o}s-R\'{e}nyi graph has $40$ vertices and a probability of $0.5$ for each independent undirected edge. The Barab\'{a}si-Albert random graph also has $40$ vertices and is formed by the preferential attachment model so that each new vertex is attached with $3$ undirected edges. See, \eg, \cite{barabasi1999emergence} for further details on the construction. For very small $r$, the MSE loss of the RBF and SVD models is similar for both graph models while the RBF approximations have a small advantage. However, the RBF model exhibits much a faster reduction of loss as the number of components grow. In particular, for the Erd\H{o}s-R\'{e}nyi model, exact reconstruction with an MSE error below $10^{-5}$ requires $18$ RBF components and $39$ SVD components, respectively. For the preferential attachment model, $8$ RBF and $37$ SVD components are needed for the exact reconstruction. 

\subsection{Molecular Distance Matrix}

In the next example, we try to approximate a distance matrix between a set of molecules in the reproducing kernel Hilbert space (RKHS) induced by a marginalized graph kernel~\cite{kashimaMarginalizedKernelsLabeled2003,tangPredictionAtomizationEnergy2019,xiangPredictingSingleSubstancePhase2021}. Here, each molecule is represented by a graph where nodes represent atoms and edges represent inter-atomic interactions. The inner product between two molecules under the graph kernel is the expectation of the path similarities between a simultaneous random walk on the two corresponding molecular graphs. Given a graph kernel $\langle \cdot, \cdot \rangle$ and a pair of molecular graphs $G$ and $G'$, the distance is computed using the geometric definition
\begin{equation}\label{eq:rkhs-distance}
d_{G, G'}^2 = \langle G - G', G - G' \rangle = \langle G, G \rangle + \langle G', G' \rangle - 2 \langle G, G' \rangle,
\end{equation}
where the subtractions in the intermediate step are to be understood as happening in the feature space.

\begin{figure}
    \includegraphics[width=\textwidth]{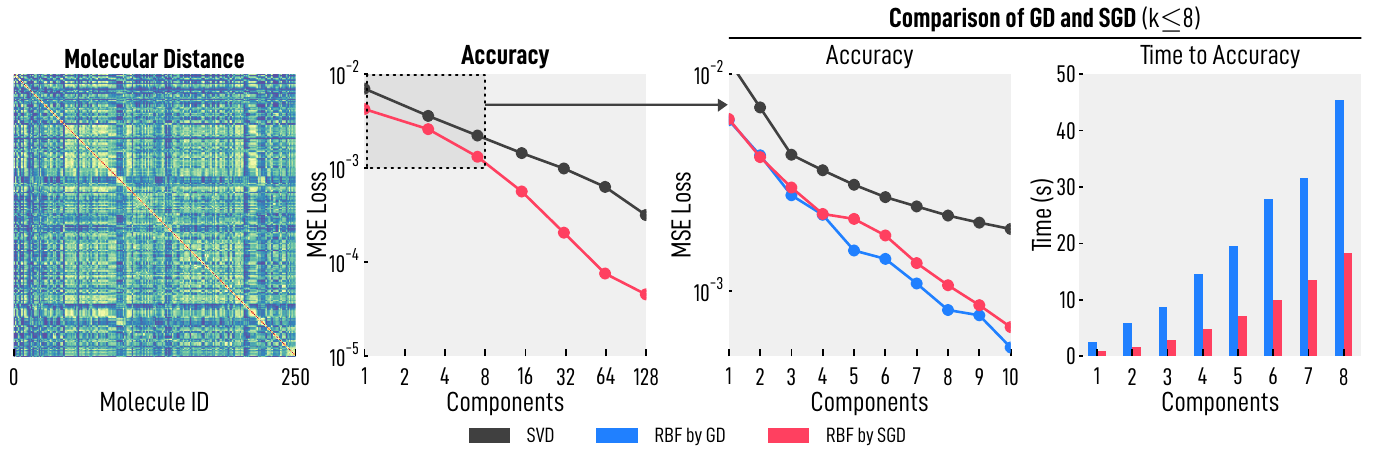}
	\caption{A comparison of RBF and SVD approximations, as well as different solution algorithms for RBF approximations, of a molecular kernel distance matrix in terms of accuracy and time consumption. (A) the molecular distance matrix; (B) construction loss of RBF and SVD approximations; (C) zoom-in comparison of the performance of full-batch gradient descent (GD) and mini-batch stochastic gradient descent (SGD) algorithms; (D) comparison of GD and SGD in terms of time-to-accuracy.
	\label{fig:molecular-distance}}
\end{figure}

The matrix in \Cref{fig:molecular-distance}(A) is computed from a random subset of 256 molecules from the QM7 dataset~\cite{blum970MillionDruglike2009}. \Cref{fig:molecular-distance}(B)/(C) show the MSE losses in log scale of the RBF approximations and the SVD approximation with the same number of ranks. To further demonstrate the advantage of the RBF approximation, in Table~\ref{table:kernel}, we collect the number of components required to achieve certain approximation accuracy, in particular, there is more than seven times memory save to reach a $0.0005$ MSE loss. 
\begin{center}
\begin{tabular}{ |c|c|c| } 
 \hline
 MSE & \# SVD ranks & \# RBF components \\ 
 \hline
 $<$ 0.01 & 3 & 1 \\ 
  0.001 & 30 & 7 \\ 
 
 0.0005 & 77 & 10  \\
 \hline
\end{tabular}
\captionof{table}{\TODO{Problem: the data here does not seem to be consistent with \Cref{fig:molecular-distance}. In that figure, for 0.001 MSE, we need rougly twice as many ranks for SVD, but not 4x as in this table.}\label{table:kernel}}
\end{center}

Green plots in \Cref{fig:molecular-distance} refer to the RBF approximation obtained by the stochastic optimization approach (discussed above at the end of Section~\ref{sec:method}) and black plots refer to the learning process based on the full gradient. In \Cref{fig:molecular-distance} (C), we can see that a non-stochastic training achieves slightly smaller MSE loss for the same parameter set-up, and \Cref{fig:molecular-distance} (D) shows that it costs us a $2.5$-$3$ times longer training time. Here, the batch size is $100$, the learning rate is $0.1$ and the number of Adam iteration steps is chosen to be $10000k$ for $k$ component approximation since this is approximately the number of iterations required for the best run of a non-stochastic training process to stabilize at some loss level. Stochastic version mini-batch size is set to be $5kn$, where $k$ is the number of components learned and $n = 256$ is the size of the matrix. Moreover, a stochastic training process can be administered significantly lighter than described, allowing to search for the RBF approximation under the limited memory resources. So, in \Cref{fig:molecular-distance} (B), we demonstrate that stochastic Adam can find an RBF approximation with $k$ being up to $128$ that significantly outperforms corresponding SVD-based approximation. For this experiment, the batch size is taken to be $50$, there is a hard limit to $50000$ iterations and mini-batch size is set to be $8n$ for every number of components $k$. We note that $64$ RBF components are observed to be enough for an almost perfect RBF approximation (with the MSE loss below 1e-4).

\section{Interpretation of the resulting decomposition}\label{sec:interpretability}

RBF components of form \eqref{eq:rbf-comp} and \eqref{eq:rbf-component-sym} can discover one-dimensional distance-related relationships because the underlying radial basis function is a distance-based positive definite kernel. As we have discussed in the previous section, the RBF approximation can reach the same $L_2$-error with the SVD approximation using significantly fewer components and parameters across a variety of natural matrix models. However, the advantages of the RBF approximation go beyond memory savings. Just like any approximation of a large data object using a few simpler and lower-dimensional components, the RBF approximation provides a means for capturing simpler underlying structures in the data. In this section, we illustrate this on a sequence of diverse applications, \eg, learning one-dimensional geometry in a three-dimensional manifold, uncovering structures in graphs, and compressing image data.

Before we proceed with the details, let us note that the applications that we consider are among the most popular and widely studied applied math problems. Consequently, each of them is associated with multiple specialized and deeply studied solution approaches. Here, our aim is \emph{not} to compare the RBF-based approach with the state-of-the-art solutions specialized to any of these applications, but rather to demonstrate the universality of the RBF approach over several diverse tasks. We hope that this can motivate the competitive promise of the RBF approximations and set a stage for further exploration. 

\textbf{One-dimensional manifold learning.} Consider an S-shaped surface
\begin{equation*}
    \left\{ \left(\sin t, 2y, \sgn(t)(\cos t - 1)\right) \mid t \in [-\frac{3}{2}\pi, \frac{3}{2}\pi], y \in [0, 2] \right\}
\end{equation*}
in a 3D space, where $t$ represents the ``time" coordinate along the curve while $y$ represents the coordinate along the ``width'' dimension of the letter S. A noisy dataset $X = (\B{x}_1, \ldots, \B{x}_{1000})$ is drawn from this surface by sampling $t$ and $y$ uniformly $1000$ times and then perturbing the points by an additive Gaussian noise $Z \sim \mathcal{N}(0, \delta \mathds{I})$.

\begin{figure}
    \includegraphics[width=\textwidth]{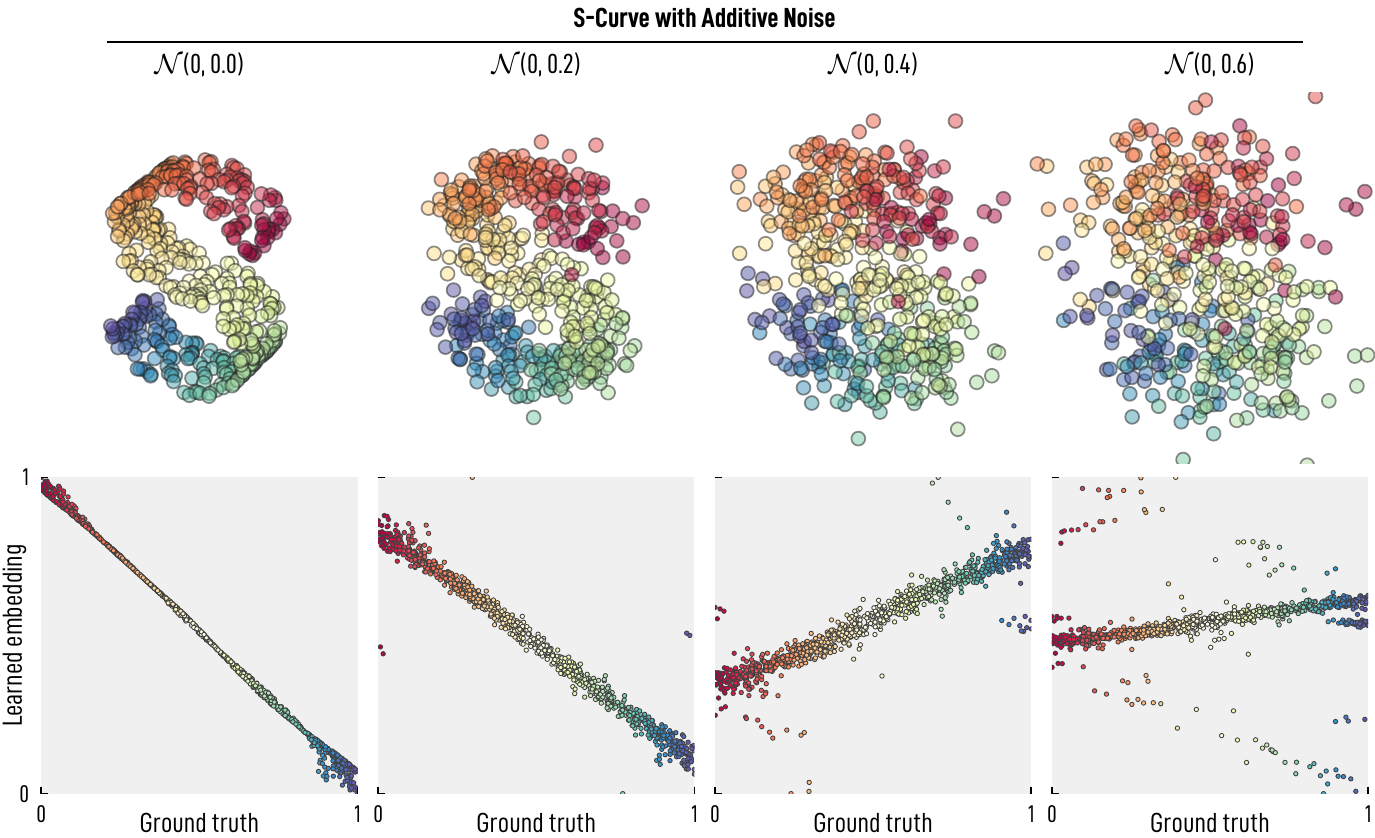}
	\caption{First row: Visualization of the S-curve dataset with various added noise levels. Second row: Linear relation between Ground truth ($\B{t}$) ``time" coordinate parametrizing given S dataset and Learned embedding ($\B{u}$) demonstrates that the increasing or decreasing order of the coordinates of $\B{u}$ can be used to uncover main one-dimensional structure of the dataset.}
    \label{fig:s-curve}
\end{figure}

The approximation target is the soft distance matrix $K$ for the dataset, namely, its $(i,j)$-th entry is defined as $k_{ij} = \exp(-\|\B{x}_i - \B{x}_j\|^2/2)$ for $i,j, = 1, \ldots, 1000$. One symmetric RBF component as per definition~\eqref{eq:rbf-component-sym} is fitted to approximate $K$, resulting in a vector $\B{u} \in \R^{1000}$ that approximately recovers the time coordinate $t$. See the relationship between the ground truth order along the curve ($\B{t} = (t_1, \ldots , t_{1000})$) and learned order ($\B{u}$) for four noise levels $\delta$ from $0$ (no noise) to $0.6$ in Figure~\ref{fig:s-curve}.

\begin{figure}
\includegraphics[width=\textwidth]{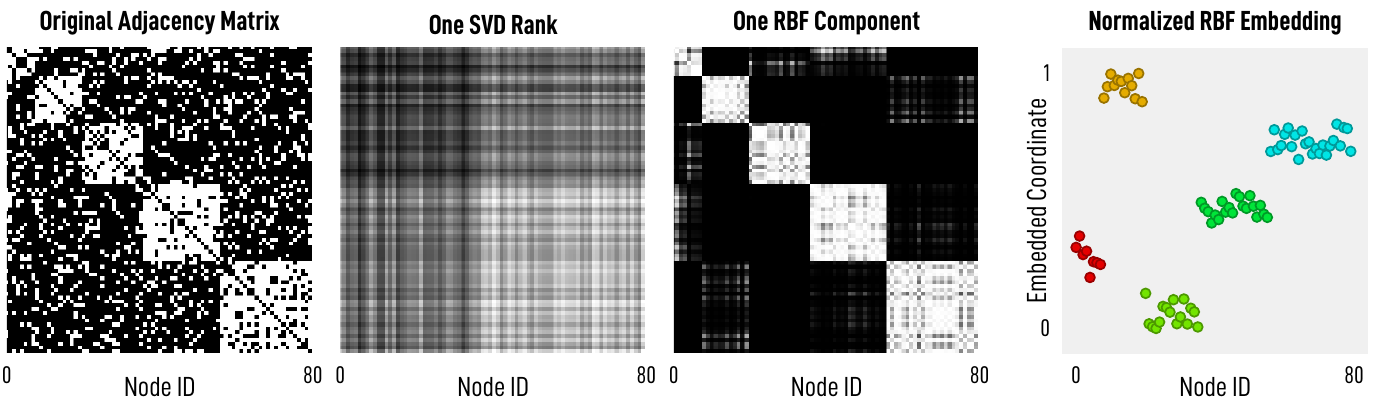}
	\caption{Left: adjacency matrix of an SBM random graph with $5$ communities with $8, 12, 16, 20$ and $24$ vertices, sampled from the model with intra-cluster probability of $0.8$ and an inter-cluster probability of $0.2$. Middle: a one component RBF approximation visually captures all $5$ communities whereas the rank one SVD-based approximation fails. Right: One-dimensional clustering of the coordinates of the RBF approximation component vector allows to recover community attribution for all the vertices of the given graph.}
    \label{fig:sbm}
\end{figure}

\textbf{Community detection.} The stochastic block model (SBM) is a random graph model with planted clusters, widely used for community detection and clustering tasks. See, \eg, \cite{abbe2017community} for a survey of modern approaches to the SBM-based clustering, as well as their limitations. One of the most standard approaches to find $k$ clusters in an SBM matrix is spectral clustering using the entries in the $k$ top leading eigenvectors. In contrast, only one RBF component can suffice to successfully detect multiple clusters. As demonstrated in \Cref{fig:sbm}, we consider an $80$-vertex graph with $5$ communities each containing $8, 12, 16, 20$ and $24$ vertices, respectively. All clusters have an intra-cluster probability of $0.8$ and an inter-cluster probability of $0.2$. Like in the previous example, we fit one symmetric RBF component~\eqref{eq:rbf-component-sym} to the adjacency matrix of the graph. A one-dimensional clustering is carried out on the learned coordinates of the component vector $u$ to assign the respective graph vertices into $5$ clusters.

\begin{figure}
    \includegraphics[width=\textwidth]{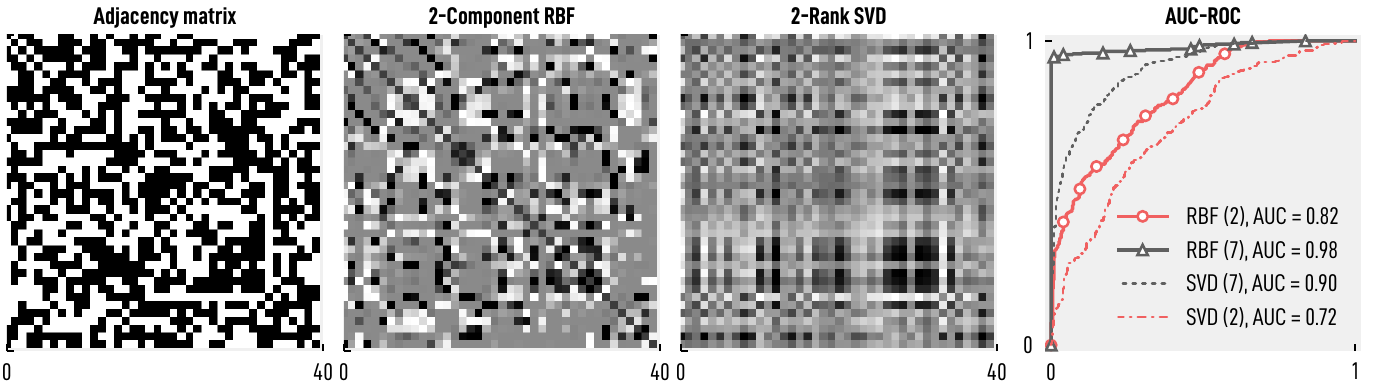}
	\caption{Left: adjacency matrix of an Erd\H{o}s-R\'{e}nyi random graph on $40$ vertices with edge probability $0.5$. Middle: a 2 component RBF approximation captures visibly more local structure comparing to the rank $2$ SVD approximation. Right: RBF-based edge prediction algorithm dominates its SVD-based counterpart for $2$-component and $7$-component approximations. Seven component RBF-based approximation allows for an almost exact edge reconstruction.} 
    \label{fig:er-inter}
\end{figure}

\textbf{Approximation of graph adjacency matrices.} We return to the example of the Erd\H{o}s-R\'{e}nyi $(40, 0.5)$ random graph considered in Section~\ref{sec:efficiency} with the observation that the $L_2$-based MSE loss is probably not the most important measure of a meaningful graph compression. Instead, what one might want to preserve in the approximation is the connectivity of the graph. In \Cref{fig:er-inter}, we can visually compare the difference in how well the RBF and SVD approximation models can capture the structural patterns in the adjacency matrix. To quantify this observation, we formalize the following \emph{edge prediction algorithm}: we threshold the real-valued entries of the approximation matrix by some cutoff value $\varepsilon \in [0,1]$ and predict an edge if the element of the approximation matrix is greater than $\varepsilon$ (thus, it is close enough to its value in the binary adjacency matrix). We confirm the advantage of the RBF approximation in retaining the true edge structure by the difference in the ROC curve and AUC score (\cite{fawcett2006introduction}) for the SVD-based and RBF-based edge prediction algorithms with $2$ and $7$ components (see Figure~\ref{fig:er-inter}, right). 

\textbf{Image data compression.} It is common to consider natural image data as inherently low-rank, this assumption makes it both theoretically and practically amenable for various denoising and compressive sensing tasks. Our preliminary experiments suggest that, unlike exactly low-rank matrices, real images can be approximated with the RBF components at least just as well as with the SVD-based components. Moreover, RBF-based approximation is even better at preserving visual acutance using the same number of parameters. Here, we compare the quality of RBF and SVD approximations on two images: 1) a baboon picture that is widely used in the image processing literature, and 2) a poster from a 2009 documentary ``Space Tourists''. For RBF approximations, we fit the images using multiple components of the general asymmetric form~\eqref{eq:rbf-comp}.

\begin{figure}
\includegraphics[width=\textwidth]{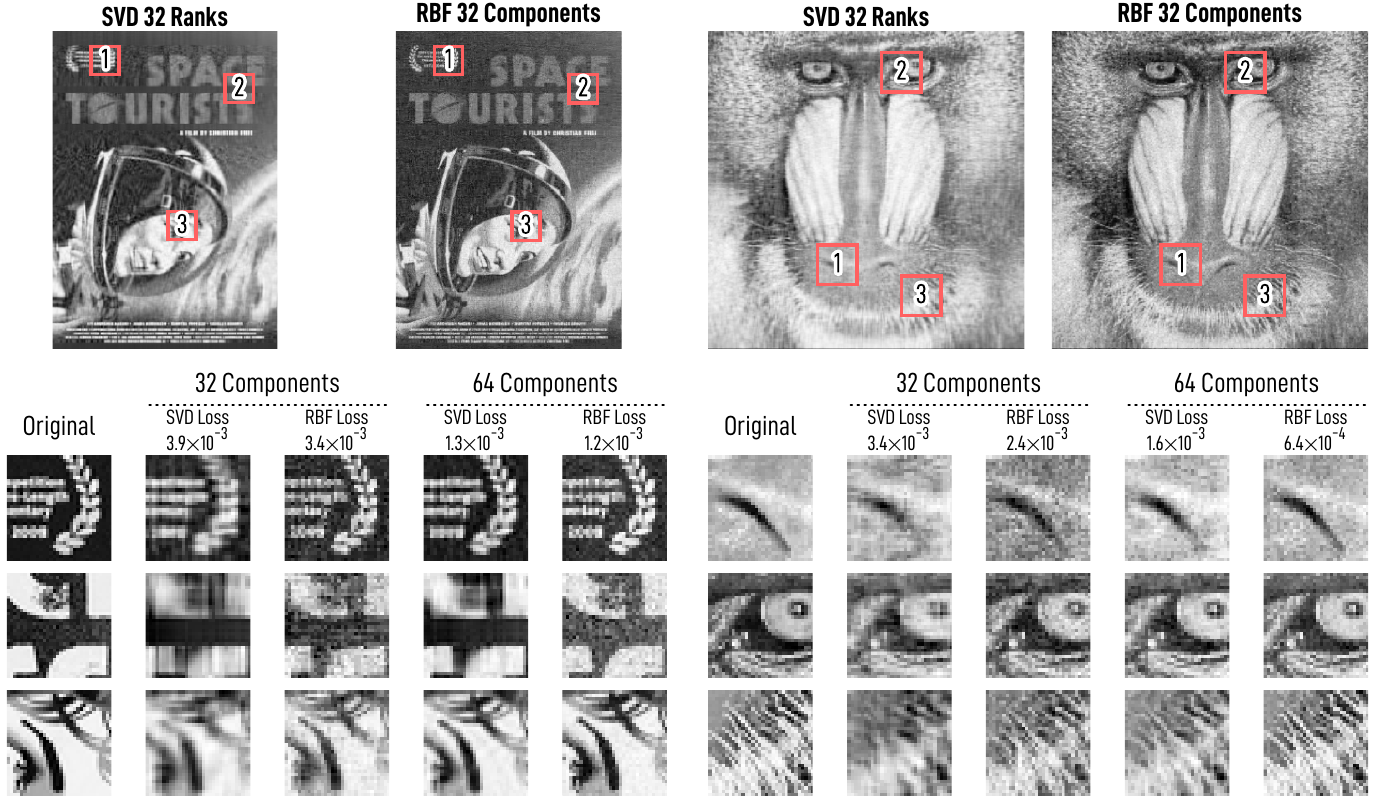}
	\caption{Visual comparison of the RBF and SVD based approximations for  ''Space tourists" poster and ''Baboon" image with $32$ and $64$ components. Low rank approximations tend to be smoother and less noisy, while RBF components produce sharper images, better preserving local features of an image, such as textures and edges.}
    \label{fig:image}
\end{figure}

For the ``baboon'' image, MSE-losses are slightly better in the RBF approximation as shown in \Cref{fig:image}, while the visual quality of the compressed image is notably higher. 
For the ``Space Tourists" poster, MSE-losses are very similar in approximations with the same number of components, however, we can also observe the advantages of the RBF compression in the visual quality both in the text and the portrait parts of the poster. It is an important follow-up direction of the current work to define the quantitative measure that captures the differences in the sharpness of the image compression better than the generic $L_2$-based MSE loss.

\section{Related Work}\label{sec:related}

The literature on the low-rank approximation of matrices is truly abundant. 

While SVD is known to give the best $L_2$ norm approximation of a given rank, numerous treatments develop efficient algorithms for computing and approximating SVD, making the computations scalable and robust, and use them to do dimension reduction, denoising, and learning from data. For a survey, see, e.g., \cite{kannan2009spectral, halko2011finding}. Let us briefly note here that slightly more sophisticated low-rank approximation tasks, such as low-rank tensor approximations, or non-negative factorizations, rapidly increase in complexity of training, as well as existence and uniqueness questions. It is common to turn to the iterative optimization algorithms to solve them (e.g., \cite{cichocki2009nonnegative}).

Beyond the low-rank approaches, special properties of the underlying data can be used to learn its approximation by some simple components. One classical example is approximation of kernel matrices. Many approaches were introduced to make kernel trick scalable, among the most prominent ones is the approximation of a kernel matrix by an scalar product of higher-dimensional vectors, namely $K(\B{x}^i, \B{x}^j) \approx \sum_k \phi(\B{x}^i)_k \phi(\B{x}^j)_k$. The function $\phi$ here can be constructed using randomly samples from the Fourier transformation of the kernel, versions of the Nystr\"{o}m method, or using polynomial approximation \cite{rahimiRandomFeaturesLargescale2007,avron2017random, wangNumericalRankRadial2018}. All these methods use positive semi-definiteness of the kernel matrix to be approximated, and sometimes its concrete functional form, and neither of which is imposed in our approach. 

The idea to use kernel functions \emph{for approximation} had several appearances in the recent literature, usually still limited to the approximation of positive semi-definite or otherwise very special matrices. For example, in \cite{duvenaudStructureDiscoveryNonparametric2013}, compositions of kernels used for nonparametric
regression. In \cite{smolik2021radial,zabran2020radial}, the authors consider approximations of functions defined on a known discrete dataset in $\mathbb{R}^d$ by linear combinations of $d$-dimensional gaussian RBF functions, as well as by more sophisticated building blocks. Although our setting is crucially different from this one, where only the coefficients are learned, but the component matrices are known by design, which results in a linear system to solve instead of a highly non-convex optimization problem.

A method that is intimately related to our current approximation approach is the Radial-Basis-Function Network. An RBF network as defined in~\cite{parkUniversalApproximationUsing1991} is a three layer neural network with RBF nonlinearities in its hidden layer. Effectively, it learns, \ie approximates, an unknown function as a linear combination of multivariate Gaussian kernel functions. RBF networks are known for the universal approximation properties \cite{parkUniversalApproximationUsing1991}. The conceptual difference between our RBF approximation with neural networks trained by a supervised learning task is that we also learn the input data. From a neural network perspective, we treat the input training data as a set of learnable parameters in addition to the weights. Moreover, in the optimization process, we are not concerned with the generalizability of the learned model because a different set of parameters are learned for \emph{each} concrete data matrix $K$. This makes our target different: we aim to make the model structure as light as possible. Clearly, an approximation exists for any data with enough overfitting, as illustrated in~\Cref{sec:efficiency}. Our primary interest is in the most compact one.

\section{Conclusion and future work}\label{sec:conclusions}
We present a novel framework for matrix data approximation based on nonlinear RBF components. We restrict the building blocks to the one-dimensional kernels, which gives us a simple way to control complexity\footnote{Moreover, multi-dimensional kernels can be approximated by one-dimensional ones, just like any other data matrix, so, considering heavier multi-dimensional components in our framework is probably redundant.}. Unlike composite kernel learning, we consider arbitrary linear combinations of kernel matrix component and deliberately do not constrain the resulting matrix to be semi-positive definite. This relaxation enhances the flexibility of the approximation and improves efficiency. We demonstrate the advantage of the proposed approximation, both in terms of memory saving and interpretability, across various datasets. We employ gradient-based method for learning the nonlinear RBF components, discuss the particulars of the learning setup and demonstrate convergence to a low-component RBF approximation. 

\bigskip

This is an initial exploratory treatment of a new method, which gives rise  to an abundance of natural follow up questions and possibilities of future extensions. For example, our current symmetric matrix fitting is bound to have ones in the diagonal. A simple correction of the functional form, such as multiplying the symmetric outer product to the RBF component matrix, can be used to approximate symmetric matrices with non-constant diagonal while keeping the number of parameters the same. More generally, changing the functional form brings in more flexibility to the method. It would be even more interesting to learn the correct functional form from the data. Another direction of a follow-up work is to formalize similar RBF-based approximations for multi-order tensor data, \eg, a multi-time series dataset, a color image, or a video clip.

Our current optimization strategy is quite generic and not specialized for a particular functional form. If the nonlinear components degenerate into standard outer products, \eg as in the case of SVD and tensor CP decomposition, then better ways are known to solve the approximation problem much faster than gradient descent based approaches. Unfortunately, the applocation of linear solver-based approaches are likely ruled out by the Gaussian nonlinearity. Nonetheless, it would be very interesting to leverage more general optimization strategies beyond gradient descent that can still benefit from some structural assumptions about the model. 

Compressed data in the form of an RBF approximation can be used as an intermediate representation in many machine learning pipelines. For example, an RBF approximation can compress neural networks models or massive training datasets. \TODO{The last sentence is a bit vague} Finally, it is a very natural future direction to further investigate particular simple data features that nonlinear approximation preserves in various applications.

\section{Acknowledgment}

This work was supported by the Laboratory Directed Research and Development Program of Lawrence Berkeley National Laboratory under U.S. Department of Energy Contract No. DE-AC02-05CH11231.

\bibliography{bibliography/yht,bibliography/er}
\bibliographystyle{unsrt}

\clearpage

\end{document}